\documentclass[letterpaper]{article} 
\usepackage{aaai23}  
\usepackage{times}  
\usepackage{helvet}  
\usepackage{courier}  
\usepackage[hyphens]{url}  
\usepackage{graphicx} 
\usepackage{natbib}  
\usepackage{caption} 
\DeclareCaptionStyle{ruled}{labelfont=normalfont,labelsep=colon,strut=off} 
\usepackage{algorithm}
\usepackage{algorithmic}
\usepackage{bm}
\usepackage{amssymb}
\usepackage{latexsym}
\usepackage{dsfont}
\usepackage{multirow}

\usepackage{color}
\newcommand{\red}[1]{\textcolor{black}{{#1}}}

\usepackage{newfloat}
\usepackage{listings}
\floatstyle{ruled}
\newfloat{listing}{tb}{lst}{}
\floatname{listing}{Listing}
\title{Hypotheses Tree Building for One-Shot Temporal Sentence Localization}

\author{Daizong Liu\textsuperscript{\rm 1,2}, Xiang Fang\textsuperscript{\rm 3}, Pan Zhou\textsuperscript{\rm 1*}, Xing Di\textsuperscript{\rm 4}, Weining Lu\textsuperscript{\rm 5}, Yu Cheng\textsuperscript{\rm 6}}
\affiliations{
\textsuperscript{\rm 1}Hubei Key Laboratory of Distributed System Security, Hubei Engineering Research Center on Big Data Security, \\ School of
Cyber Science and Engineering, Huazhong University of Science and Technology\\
\textsuperscript{\rm 2}Peking University
\textsuperscript{\rm 3}Nanyang Technological University \\
\textsuperscript{\rm 4}ProtagoLabs Inc \
\textsuperscript{\rm 5}Tsinghua University \
\textsuperscript{\rm 6}Microsoft Research
\\
dzliu@hust.edu.cn, xfang9508@gmail.com, panzhou@hust.edu.cn, xing.di@protagolabs.com, \\ luwn@tsinghua.edu.cn, yu.cheng@microsoft.com
}

\begin{document}

\maketitle

\begin{abstract}
Given an untrimmed video, temporal sentence localization (TSL) aims to localize a specific segment according to a given sentence query.
Though respectable works have made decent achievements in this task, they severely rely on dense video frame annotations, which require a tremendous amount of human effort to collect.
In this paper, we target another more practical and challenging setting: one-shot temporal sentence localization (one-shot TSL), 
which learns to
retrieve the query information among the entire video with only one annotated frame.
Particularly, we propose an effective and novel tree-structure baseline for one-shot TSL, called Multiple Hypotheses Segment Tree (MHST), to capture the query-aware discriminative frame-wise information under the insufficient annotations.
Each video frame is taken as the leaf-node, and the adjacent frames sharing the same visual-linguistic semantics will be merged into the upper non-leaf node for tree building.
At last, each root node is an individual segment hypothesis containing the consecutive frames of its leaf-nodes.
During the tree construction, we also introduce a pruning strategy to eliminate the interference of query-irrelevant nodes.
With our designed self-supervised loss functions, our MHST is able to generate high-quality segment hypotheses for ranking and selection with the query.
Experiments on two challenging datasets demonstrate that MHST achieves competitive performance compared to existing methods.
\end{abstract}

\section{Introduction}
\begin{figure}[t]
\centering
\includegraphics[width=0.48\textwidth]{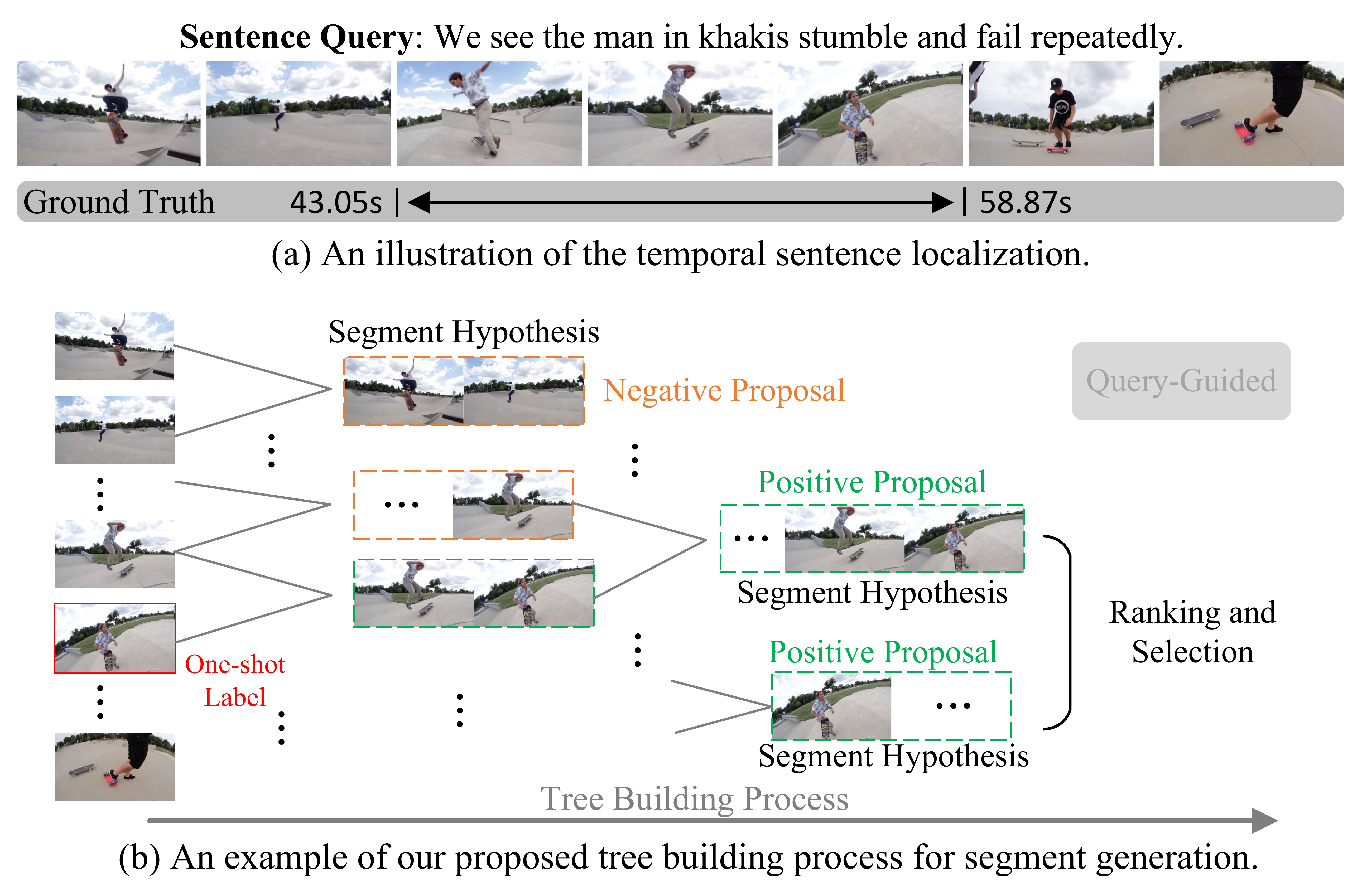}
\caption{(a) The example of the temporal sentence localization (TSL) task. (b) Illustration of the one-shot TSL setting and our proposed tree-structure method.}
\label{fig:intro}
\vspace{-10pt}
\end{figure}

Temporal sentence localization (TSL) \cite{anne2017localizing,gao2017tall} has drawn increasing attention in recent years, which aims to retrieve a temporal video segment that semantically corresponds to a given sentence query.
An illustrative example of TSL is shown in Figure~\ref{fig:intro} (a). Clearly, this task involves both computer vision and natural language processing techniques for multi-modal encoding and cross-modal reasoning, to correctly locate the segment boundaries from an untrimmed video.

Most previous TSL works \cite{liu2018cross,ge2019mac,yuan2019semantic,zhang2019man,zhang2020span,chenrethinking,yuan2019find,liu2020jointly,liu2020reasoning,liu2021adaptive,liu2021progressively,liu2022exploring,liu2022memory} proposed for this task are under fully-supervised setting, where each frame is manually labeled as the query-relevant or query-irrelevant frame.
In spite of their great advances, these methods severely rely on abundant video-query annotations, which is labor-intensive and time-consuming to collect in real-world scenarios.
To alleviate this problem, some recent works explore a weakly-supervised setting \cite{mithun2019weakly,lin2020weakly,zhang2020counterfactual,liu2022unsupervised,chen2020look,song2020weakly} with only the video-query correspondence rather than the dense frame annotations.
However, their performances are validated to be less satisfied with such weak supervision.

In this paper, we \red{aim to tackle a more practical and challenging scenario for TSL task, \textit{i.e.}, one-shot TSL, which only has the label of one query-relevant frame in each video.
This one-shot TSL setting is first proposed by ViGA \cite{cui2022video}, however, the performance of this model is severely underestimated by current limited designs.}
We find that previous supervised TSL methods \red{also} can not be directly applicable to this \red{challenging} setting due to the following limitations:
1) Firstly, most existing works generally pre-define multiple segment proposals and utilize the dense annotations to score these proposals for ranking and selection. However, one-shot TSL has only one frame label in each video, lacking sufficient knowledge to build and score the segment proposals.
2) Secondly, there are many unlabeled frames that may be either irrelevant to the query or the labeled frame within each video.
These frames may bring confounding between the frame label and irrelevant visual features.
Therefore, how to discriminate their frame-wise representations for precise segment boundary estimation is also a challenging issue.
3) Thirdly, with limited supervision signals, traditional supervised loss functions are ineffective enough to train the one-shot TSL models. Thus, an appropriate framework and the training strategy should be well-designed for this one-shot setting.

To this end, we propose \red{a novel and effective} baseline model for one-shot TSL task, called Multiple Hypotheses Segment Tree (MHST), which adopts a tree structure to generate learnable segment hypothesis by merging the adjacent frames sharing the same visual-linguistic semantics with the query and the labeled frame.
As shown in Figure~\ref{fig:intro} (b), the tree building process is taken as the segment hypothesis construction process.
Specifically, we start from treating video frames as the initial leaf-nodes, and then merge the adjacent nodes to the upper non-leaf nodes based on the visual relevance of themselves and their linguistic relevance with the query. During the tree building, we further introduce a tree pruning strategy to selectively weaken the impact of the query-irrelevant nodes. The final hypotheses tree will have several root nodes, which represent corresponding segment hypotheses constructed by the frames of its contained leaf-nodes.
At last, we take the segments containing the labeled frame as positive samples and the others as the negative ones. We devise several self-supervised training losses to jointly learn discriminative frame-wise representations for accurately segment hypotheses scoring and selection.
During the inference stage, we directly choose the segment hypothesis with the highest confidence score as the final prediction.
To sum up, our main contributions are as follows:
\begin{itemize}
    \item \red{In this paper, we investigate the structure limitations of previous TSL methods on the challenging one-shot setting.} Specifically, we devise a new tree-structure framework for one-shot TSL, called MHST, to construct query-related segment hypothesis based on the solely one labeled frame.
    \item As for the hypotheses tree building, we propose to merge the adjacent nodes into their upper node based on both visual and linguistic relevances. We also introduce a tree pruning strategy to filter out the query-irrelevant nodes. During the training, we apply the self-supervised losses to learn the model under limited annotations.
    \item We conduct comprehensive experiments on two challenging datasets (ActivityNet Captions and Charades-STA). The results demonstrate the effectiveness of our proposed method, where MHST achieves decent results and outperforms most fully-supervised methods.
\end{itemize}

\section{Related Work}
\noindent \textbf{Temporal sentence localization.}
Most of the existing TSL methods refer to fully-supervised setting where all video-query pairs are annotated in details, including corresponding segment boundaries.
Therefore, the main challenge in such setting is how to align multi-modal features well to predict precise boundary.
Some works \cite{liu2022reducing,liu2022skimming,liu2022learning,chen2018temporally,liu2018cross,ge2019mac,guo2022hybird,zhang2019man,liu2021context,liu2022exploringa,fang2021animc,fang2022hierarchical,fang2022multi,liu2022few} integrate sentence information with each fine-grained video clip unit, and predict the scores of candidate segments by gradually merging the fusion feature sequence over time. 
Although these methods achieve good performances, they severely rely on the quality of the segment proposals and are time-consuming. Without using proposals, some latest methods \cite{nan2021interventional,zhang2020span,chenrethinking,yuan2019find} are proposed to leverage the interaction between video and sentence to directly predict the starting and ending frames. 
However, the above methods heavily rely on the datasets that require numerous manually labelled annotations for training.

To ease the human labelling efforts, several recent works \cite{mithun2019weakly,chen2020look,song2020weakly,lin2020weakly,zhang2020counterfactual} consider a weakly-supervised setting which only access the information of matched video-query pairs without accurate segment boundaries.
However, their performance is less satisfied with such weak supervision.
Considering that we are more likely to have a limited annotation budget
rather than full annotation or no annotation in practice, 
\red{\cite{cui2022video} introduce a new practical setting for TSL task, \textit{i.e.,} solely labeling one query-relevant frame in each video. However, the performance of this model is
severely underestimated by current limited designs.}

\noindent \textbf{Multiple hypotheses construction.}
Multiple hypotheses strategies are firstly used in the field of object tracking \cite{blackman1999design,blackman2004multiple} in videos.
Their widely used hypotheses tracking \cite{cox1996efficient} algorithm originally evaluates its usefulness in the context of visual tracking and motion correspondence.
To further improve the tracking quality, multiple hypotheses tracking in \cite{kim2015multiple} propose a tracking tree with a scoring function to prune the hypothesis space efficiently and accurately which is suited to current visual tracking context.
In our method, we adapt the multiple hypotheses strategy to the video segment construction scenario, where propagation between the adjacent frames is newly determined by their visual-linguistic relevance instead of the unreliable appearance similarity. 
We build such multiple hypotheses tree to generate query-related segment proposals within each video by grouping the query-aware semantically closed adjacent frames, and measure their scores for selection.
Different from \cite{li2022end}, we propose newly designed self-supervised losses for TSL task.

\begin{figure*}[t!]
\centering
\includegraphics[width=0.98\textwidth]{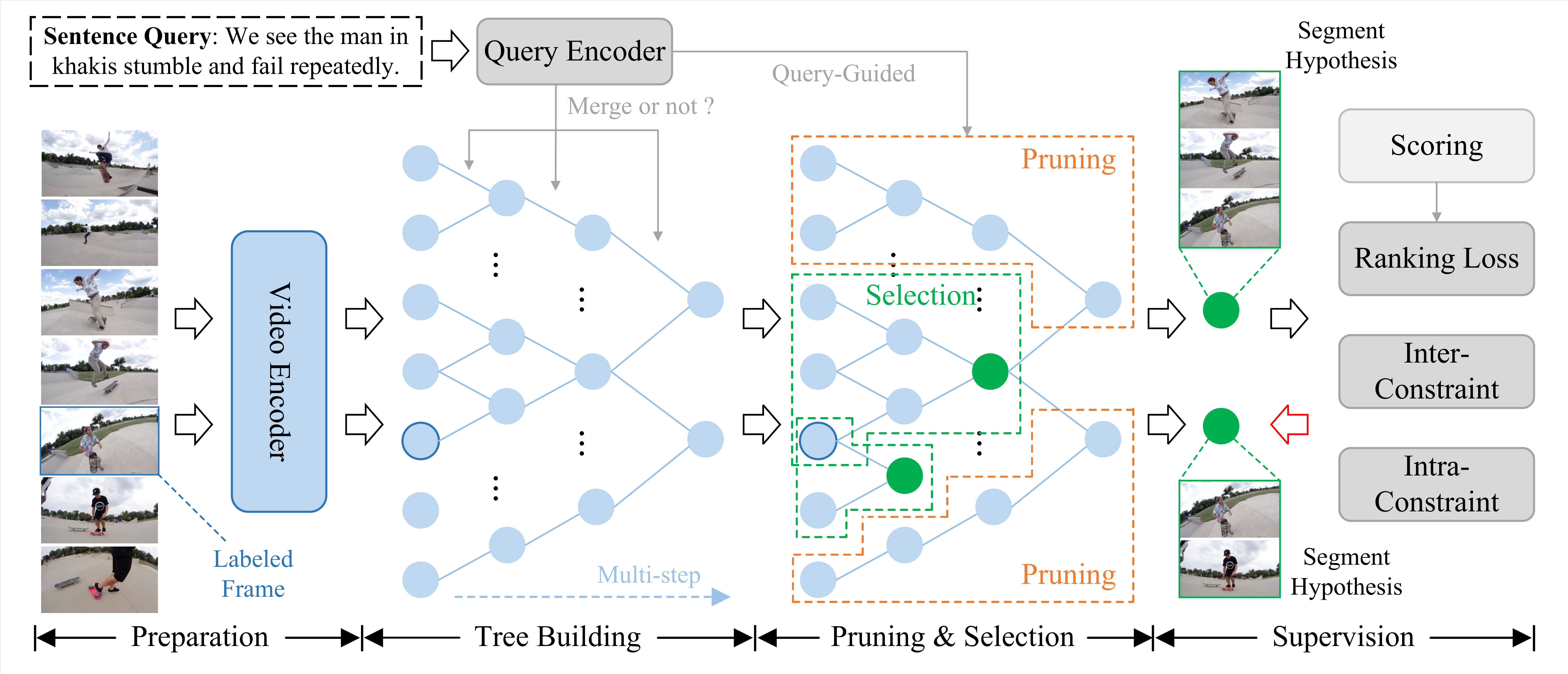}
\caption{An overview of the proposed architecture for the one-shot TSL task. Given the video-query input, we first take each video frame as leaf-node, and then iteratively build the hypotheses segment tree by merging the nodes sharing the same visual-linguistic semantics. During the tree building, we also employ a tree pruning strategy to remove and down-weight the query-irrelevant nodes. After that, we take the final tree root nodes as the segment hypotheses. At last, we utilize several self-supervised losses to score and rank these segments for learning more discriminative representations.}
\label{fig:pipeline}
\vspace{-10pt}
\end{figure*}

\section{The Proposed Model}
\subsection{Overview}
\noindent \textbf{Problem definition.}
Given an untrimmed video $V=\{v_i\}_{i=1}^{N_v}$ and the natural language query $Q=\{q_i\}_{i=1}^{N_q}$, where $N_v$ and $N_q$ are the number of frames and words, the temporal sentence localization (TSL) task aims to localize the query-described activity segment from the video. 
Particularly, in one-shot TSL setting, there is solely one matched frame annotated with the query in video $V$.
Therefore, it is more challenging than previous TSL works to predict the accurate segment with the limited supervision signals.

\noindent \textbf{Pipeline.}
In this section, considering the few supervision in one-shot TSL task, we propose a novel pipeline named Multiple Hypotheses Segment Tree (MHST) to construct multiple segment hypotheses that are visual-related and linguistic-related to the labeled frame and the query.
Specifically, as shown in Figure~\ref{fig:pipeline}, MHST contains four main steps:
\textit{Firstly}, given a video-query pair, we extract frame-wise features via a video encoder and extract sentence-level features via a query encoder for preparation.
\textit{Secondly}, we build the hypotheses tree to discriminate frame-wise information for segment construction. At the beginning, each video frame is taken as a leaf-node. Then, the adjacent nodes will be iteratively
merged into an upper node based on their visual-relevance and the query-relevance.
\textit{Thirdly}, to avoid the tree explosion and the negative impact of unnecessary nodes, we further introduce a tree pruning strategy to remove and down-weight the nodes conditioned on the query.
\textit{At last}, the final generated root nodes are the segment hypotheses constructed by the frames of their contained leaf-nodes. We develop a scoring head to rank these segments and design two self-supervised losses to train the whole hypotheses tree for discriminating the frame-wise representation.

\subsection{Preparation}
Before building the hypotheses tree, we first extract both video and query features for preparation.

\noindent \textbf{Video encoding.}
Following previous works \cite{zhang2019cross,zhang2020span}, given the video $V$, we first extract its frame-wise features by a pre-trained C3D network \cite{tran2015learning}, and then employ a multi-head self-attention \cite{vaswani2017attention} module to capture the long-range dependencies among video frames. We denote the extracted video features as $\bm{V}=\{\bm{v}_i\}_{i=1}^{N_v} \in \mathbb{R}^{N_v \times d}$, where $d$ is feature dimension.

\noindent \textbf{Query encoding.}
Given the query $Q$, we first utilize the Glove \cite{pennington2014glove} model to embed each word into dense vector, and then employ Bi-GRU \cite{chung2014empirical} to encode its sequential information. The final sentence-level feature $\bm{q} \in \mathbb{R}^{d}$ can be obtained by concatenating the last hidden unit outputs in Bi-GRU.

\subsection{Hypotheses Tree Building}
Here, we will illustrate how to construct the hypotheses tree for generating the possible query-related segments.
Given the frame features $\{\bm{v}_i\}_{i=1}^{N_v}$, we treat each frame as the initial leaf-node, and iteratively merge the adjacent frames as the upper non-leaf node if they share the same visual-linguistic semantics. We repeat this merging process until there is none matched adjacent nodes and take the final node as the root node. Therefore, the leaf-nodes contained in the root node are the consecutive frames for constructing the possible query-related segment. To generate multiple root nodes in once tree building process, we try to measure and merge each pair of adjacent leaf-nodes at the beginning.

Specifically, we take two leaf-nodes $\bm{v}_i,\bm{v}_j$ as an example to illustrate the node-wise merging process.
To determine whether the node pair $\bm{v}_i,\bm{v}_j$ should be merged into the same segment (or upper node), we calculate both the linguistic relevance between each node pair with the language query and the visual relevance between the nodes as the judgments.
For the linguistic relevance calculation, we first compute the query-guided information relation between the query feature and each frame feature as follows:
\begin{equation}
\label{eq:2}
    r^{qv}_{v_i} = sigmoid((\bm{W}_1 \bm{v}_i) (\bm{W}_2\bm{q})^{\top}),
\end{equation}
\begin{equation}
    r^{qv}_{v_j} = sigmoid((\bm{W}_1 \bm{v}_j) (\bm{W}_2\bm{q})^{\top}),
\end{equation}
where $\bm{W}_1,\bm{W}_2$ are the projection matrices. Then, we measure the query-aware difference between the node pair via $r^{qv}=|r^{qv}_{v_i}-r^{qv}_{v_j}|$.
For the visual relevance calculation, we directly compute the appearance relation between two frame features via a cosine similarity function as follows:
\begin{equation}
    r^{vv} = \frac{\bm{v}_i \bm{v}_j^{\top}}{||\bm{v}_i||_2||\bm{v}_j||_2}.
\end{equation}
Therefore, we can utilize both linguistic and visual relevances as scores to evaluate the semantic difference between the adjacent node pair $\bm{v}_i,\bm{v}_j$ by:
\begin{equation}
\label{eq:1}
    r = \lambda_1 r^{qv} + \lambda_2 r^{vv},
\end{equation}
where $\lambda_1,\lambda_2$ are used to control the balance.

With the relevant difference value $r$, we can rank all node pairs in each step and pick out the top $\alpha$, which is a hyperparameter to be set as a percentage. For the top $\alpha$ node pair $\bm{v}_i,\bm{v}_j$, we merge them into a new ancestor non-leaf node:
\begin{equation}
\label{eq:4}
    \bm{v}^{new}_{i,j} = \bm{W}_3 \bm{v}_i + \bm{W}_3 \bm{v}_j + \bm{b},
\end{equation}
where $\bm{W}_3,\bm{b}$ are the learnable weights.
During the hypotheses tree construction, we repeat this node-merging process until there is no-relevant node pairs.

\subsection{Hypotheses Tree Pruning}
During the building process of the hypotheses tree, not all the non-leaf nodes in the tree branches are closely related to the sentence query.
Therefore, we have to take a pruning step to remove these non-leaf nodes and their descendant non-leaf nodes.
In other words, we need to determine the mostly likely query-relevant non-leaf nodes for accurately constructing most query-relevant video segment.
To this end, we first calculate the semantic relevance $r$ for each non-lead node and the query following the Equation (\ref{eq:1}) and then remove the nodes that are with semantic relevance less than a hyperparameter $\tau$.
To reserve enough frame nodes (leaf-nodes) for other hypotheses construction, we do not remove their leaf-nodes. Instead, we down-weight the contribution of these lead-nodes to the final segment as follows:
\begin{equation}
\label{eq:3}
    (\bm{v}_i)' = \lambda_{\tau} r_{v_i}^{qv} \bm{v}_i,
\end{equation}
where $r_{v_i}^{qv}$ is the linguistic relevance in Equation (\ref{eq:2}) to measure the similarity between frame node and query. In particular, we apply this down-weighting process to all leaf-nodes, where $\lambda_{\tau}$ of the lead-nodes in the removed non-leaf nodes are set to 0.5, $\lambda_{\tau}$ of other leaf-nodes are set to 1.0.

We take a $L$-scan pruning method to prune the disturbing non-leaf nodes gradually instead of pruning the whole tree.
In particular, for every $L$ step, we apply the pruning process to remove the query-irrelevant non-leaf nodes in current step. Then, we track and remove their descendant non-leaf nodes in previous $L-1$ step. Finally, we down-weight their leaf-nodes via Equation (\ref{eq:3}).

\subsection{Segment Hypothesis Selection and Training}
After the hypotheses tree building and pruning, we can get the final segment hypotheses tree in which each root node represents a segment hypothesis containing the consecutive frames of its descendant leaf-nodes.
We directly select the root nodes that contain the ground-truth annotated leaf-node as the positive hypotheses, and denote their corresponding segment proposals as $P=\{p_i\}_{i=1}^{N_p}$ where $N_p$ is the proposal number.
To generate confidence scores $S=\{s_i\}_{i=1}^{N_p}$ for these segment proposals, we feed their merged node-wise features (obtained by Equation (\ref{eq:4})) into a fully connected layer with further sigmoid function.
Then, we apply the selection algorithm that considers the confidence score to select the top-$K$ proposals $P^K=\{p^K_i\}_{i=1}^{K}$ and give their corresponding confidence scores as $S^K=\{s^K_i\}_{i=1}^{K}$.

\noindent \textbf{Training.}
In order to learn and correct the above confidence scores, we apply a ranking loss based on a reward policy.
Specifically, we define the reward $R^K_i$ for the proposal $p^K_i$ with a reward function to encourage proposals with higher linguistic relevance $r^{qv}$.
Then the strategy of policy gradient is used to correct the scores. Note that the confidence scores are normalized by a softmax layer, which is an extremely important operation to highlight the semantically matching proposals and weaken the mismatched ones. The ranking loss can be formulated by:
\begin{equation}
\label{eq:rank}
    \mathcal{L}_{rank} = \sum_{i=1}^K -R^K_i log(\frac{exp(s_i^K)}{\sum_{j=1}^K exp(s_j^K)}).
\end{equation}

Beside, to further assist the node-wise representation learning during the hypotheses tree construction, we additionally develop an \textit{inter-constrain} and an \textit{intra-constrain} losses to train the tree-structure framework.
As for the \textit{inter-constrain} loss, we take the matched video-sentence pairs as positive samples and the unmatched video-sentence pairs as negative samples, and use the weighted binary cross-entropy loss to supervise the query-relevance $r^{qv}$ of the top-$K$ hypothesis segments as:
\begin{equation}
\label{eq:inter}
    \mathcal{L}_{inter} = \sum_{j=1}^J\sum_{i=1}^K (-y_jlog(r^{qv}_i)-(1-y_j)log(1-r^{qv}_i))),
\end{equation}
where $J$ is the number of all video-sentence pairs, $y_j$ the label of the $j$-th sample that equals 1 for the matched video-sentence pairs and 0 for the unmatched video-sentence pairs.
As for the \textit{intra-constrain} loss, within the same video, we take the top-$K$ segment proposals as positive samples and define the random segments unoverlaped with the top-$K$ segment proposals as negative samples. We adopt a hinge loss to supervise their semantics as:
\begin{equation}
\label{eq:intra}
    \mathcal{L}_{intra} = \sum_{i=1}^K max(0,\beta-r^{qv}_i+\bar{r}^{qv}_i),
\end{equation}
where $\bar{r}^{qv}_i$ denotes the query-relevance of negative samples. During the training, we utilize three fixed weights to balance the values of above three losses to joint learn the model.

\noindent \textbf{Testing.}
During the testing, we directly choose the root node from the constructed tree with the maximum confidence score, and then generate corresponding segment as the final prediction.

\section{Experiments}
\subsection{Dataset}
\noindent \textbf{ActivityNet Captions.}
This dataset is built from ActivityNet v1.3 dataset \cite{caba2015activitynet} for dense video captioning. It contains 20000 YouTube videos with 100000 queries. On average, videos are about 120 seconds and queries are about 13.5 words. We follow the public split of the dataset that contains a training set and two validation sets val 1 and val 2. Following common settings, we use val 1 as our validation set and use val 2 as our testing sets.

\noindent \textbf{Charades-STA.}
This dataset is built from the Charades \cite{sigurdsson2016hollywood} dataset and transformed into temporal sentence localization task by \cite{gao2017tall}. It contains 16128 video-sentence pairs with 12408 pairs used for training and 3720 for testing. The videos are about 30 seconds on average. The annotations are generated by sentence decomposition and keyword matching with manually check.

\begin{table}[t!]
\small
\begin{center}
\scalebox{1.0}{
\setlength{\tabcolsep}{1.4mm}{
\begin{tabular}{l|c|cccc}
\hline
\multirow{3}*{Methods} & \multirow{3}*{Type} & \multicolumn{4}{c}{ActivityNet Captions} \\ \cline{3-6}
~ & ~ & R@1 & R@1 & R@5 & R@5\\
~ & ~ & IoU=0.3 & IoU=0.5 & IoU=0.3 & IoU=0.5\\
\hline\hline
Random & FS & 18.64 & 7.73 & 52.78 & 29.49\\
VSA-RNN & FS & 39.28 & 24.43 & 70.84 & 55.52\\
VSA-STV & FS & 41.71 & 24.01 & 71.05 & 56.62\\
CTRL & FS & - & 29.01 & - & 59.17\\
TGN & FS & 43.81 & 27.93 & 54.56 & 44.20 \\
2D-TAN & FS & 59.45 & 44.51 & 85.53 & 77.13 \\
IVG-DCL & FS & 63.22 & 43.84 & - & - \\
DRN & FS & - & 45.45 & - & \textbf{77.97}\\
\hline\hline
CTF & WS & 44.30 & 23.60 & - & -\\
ICVC & WS & 46.62 & 29.52 & 80.92 & 66.61 \\
MARN & WS & 47.01 & 29.95 & 72.02 & 57.49\\
SCN & WS & 47.23 & 29.22 & 71.45 & 55.69\\
LCNet & WS & 48.49 & 26.33 & 82.51 & 62.66\\
CCL & WS & 50.12 & 31.07 & 77.36 & 61.29 \\
VCA & WS & 50.45 & 31.00 & 71.79 & 53.83 \\
WSTAN & WS & 52.45 & 30.01 & 79.38 & 63.42\\
CRM & WS & 55.26 & 32.19 & - & - \\
\hline\hline
\red{ViGA} & \red{OS} & \red{59.61} & \red{35.79} & \red{-} & \red{-} \\
\textbf{MHST} & OS & \textbf{64.34} & \textbf{45.68} & \textbf{86.92} & 77.75 \\
\hline
\end{tabular}}}
\end{center}
\caption{Performance comparisons on ActivityNet Captions dataset, where FS: fully-supervised setting, WS: weakly-supervised setting, and OS: one-shot setting.}
\vspace{-10pt}
\label{tab:activity}
\end{table}

\subsection{Experimental Settings}
\noindent \textbf{Evaluation metric.}
We adopt “R@n, IoU=m” as the evaluation metrics, following previous works \cite{gao2017tall,liu2018attentive,zhang2020span}. The “R@n, IoU=m” denotes the percentage of language queries having at least one result whose Intersection over Union (IoU) with ground truth is larger than m in top-n retrieved segment hypotheses. 

\noindent \textbf{Implementation details.}
In order to make a fair comparison with previous works, we utilize the pre-trained C3D \cite{tran2015learning} model to extract video features and employ the Glove model \cite{pennington2014glove} to obtain word embeddings. As some videos are too long, we set the length of video feature sequences to 128 for Charades-STA and 256 for ActivityNet Captions, respectively. We fix the query length to 10 in Charades-STA and 20 in ActivityNet Captions. The feature dimension $d$ is set to 512. For the hyper-parameters, we set the percentage $\alpha$ to 60\%, and set the pruning threshold $\tau$ as 0.7. The balanced weights $\lambda_1,\lambda_2$ are set to 1.0,1.0. The step $L$ in L-scan pruning is set to 3.
During training, the learning rate is by default 0.00005, and decays by a factor of 10 for every 35 epochs. The batch size is 1 and the maximum training epoch is 100. For one-shot setting, we randomly selection one labeled frame from the ground-truth as the annotation for each video.
All the experiments are implemented by PyTorch.

\begin{table}[t!]
\small
\begin{center}
\scalebox{1.0}{
\setlength{\tabcolsep}{1.4mm}{
\begin{tabular}{l|c|cccc}
\hline
\multirow{3}*{Methods} & \multirow{3}*{Type} & \multicolumn{4}{|c}{Charades-STA}\\ \cline{3-6}
~ & ~ & R@1 & R@1 & R@5 & R@5 \\
~ & ~ & IoU=0.5 & IoU=0.7 & IoU=0.5 & IoU=0.7\\
\hline\hline
Random & FS & 8.51 & 3.03 & 37.12 & 14.06 \\
VSA-RNN & FS & 10.50 & 4.32 & 48.43 & 20.21 \\
VSA-STV & FS & 16.91 & 5.81 & 53.89 & 23.58 \\
CTRL & FS & 23.62 & 8.89 & 58.92 & 29.52\\
2D-TAN & FS & 39.81 & 23.25 & 79.33 & 52.15\\
DRN & FS & 45.40 & 26.40 & 88.01 & 55.38 \\
IVG-DCL & FS & \textbf{50.24} & 32.88 & - & - \\
\hline\hline
SCN & WS & 23.58 & 9.97 & 71.80 & 38.87 \\
CTF & WS & 27.30 & 12.90 & - & - \\
WSTAN & WS & 29.35 & 12.28 & 76.13 & 41.53 \\
ICVC & WS & 31.02 & 16.53 & 77.53 & 41.91 \\
MARN & WS & 31.94 & 14.18 & 70.00 & 37.40 \\
CCL & WS & 33.21 & 15.68 & 73.50 & 41.87 \\
CRM & WS & 34.76 & 16.37 & - & - \\
VCA & WS & 38.13 & 19.57 & 78.75 & 37.75 \\
LCNet & WS & 39.19 & 18.17 & 80.56 & 45.24 \\
\hline\hline
\red{ViGA} & \red{OS} &  \red{35.11} & \red{15.11} & \red{-} & \red{-} \\
\textbf{MHST} & OS & 49.62 & \textbf{34.48} & \textbf{89.29} & \textbf{57.50} \\
\hline
\end{tabular}}}
\end{center}
\caption{Performance comparisons on Charades-STA dataset, where FS: fully-supervised setting, WS: weakly-supervised setting, and OS: one-shot setting.}
\vspace{-10pt}
\label{tab:charades}
\end{table}

\subsection{Comparison with State-of-the-Art}
\noindent \textbf{Compared methods.}
We compare MHST with the state-of-the-art TSL methods grouped into three categories:
1) Fully-supervised setting: Random Selection \cite{gidaris2018unsupervised}, VSA-RNN and VSA-STV \cite{gao2017tall}, CTRL \cite{gao2017tall}, TGN \cite{chen2018temporally}, 2D-TAN \cite{zhang2019learning}, IVG-DCL \cite{nan2021interventional}, DRN \cite{zeng2020dense}.
2) Weakly-supervised setting:
CTF \cite{chen2020look},
ICVC \cite{chen2022explore},
MARN \cite{song2020weakly},
SCN \cite{lin2020weakly},
LCNet \cite{yang2021local},
CCL \cite{zhang2020counterfactual},
VCA \cite{wang2021visual},
WSTAN \cite{wang2021weakly},
CRM \cite{huang2021cross}.
\red{3) One-shot setting: ViGA \cite{cui2022video}.}

\noindent \textbf{Comparison and analysis.}
As shown in Table~\ref{tab:activity} and \ref{tab:charades}, we compare our method with existing works on both ActivityNet Captions and Charades-STA datasets, respectively. From the tables, we have the following findings:
\begin{itemize}
    \item Firstly, compared to the weakly-supervised methods, our method outperforms them by a large margin. Specifically, on ActivityNet Captions dataset, compared to the SOTA method CRM, we bring the improvement of 9.08 and 13.49 in terms of R@1, IoU=0.3 and R@1, IoU=0.5. On Charades-STA dataset, we also outperform the SOTA method LCNet by 10.43, 16.31, 8.73 and 12.26 on all metrices. \textit{Note that}, \red{the} one-shot setting costs similar human labors as the weakly-supervised setting since the latter also requests the annotators to determine the matched video-query pair with at least one query-relevant frame. However, our performance is much better than the weakly-supervised one, demonstrating the effectiveness of the proposed tree-structure framework.
    \item Secondly, compared to the fully-supervised methods, our one-shot setting has much less annotations (one labeled frame vs. dense video annotations) for model training. However, as shown in the tables, our method achieves very competitive performances, which are even better than the previous SOTA methods. Specifically, on ActivityNet Captions dataset, compared to DRN, we outperform it by 0.23 in R@1, IoU=0.5. On Charades-STA dataset, compared to IVG-DCL, we bring improvement of 1.60 in R@1, IoU=0.7. This demonstrates that our framework is robust to the weak annotations. Although we have much less annotations than the fully-supervised works, our well-designed tree-structure framework can group the query- and labeled frame-related adjacent frames into the same segment under the supervision of the proposed effective self-supervised losses. 
    \item \red{Thirdly, our performance outperforms ViGA method a lot, demonstrating that our tree-structure pipeline is more effective and reasonable for the one-shot TSL setting.}
\end{itemize}

\subsection{Ablation Study}
In this section, we conduct ablation study to validate the effectiveness of each components on ActivityNet Captions.

\begin{table}[t!]
\small
\begin{center}
\scalebox{1.0}{
\setlength{\tabcolsep}{0.6mm}{
\begin{tabular}{ccc|cccc}
\hline
Segment & Tree & Self- & R@1 & R@1 & R@5 & R@5\\
Tree & Pruning & supervision & IoU=0.3 & IoU=0.5 & IoU=0.3 & IoU=0.5\\
\hline\hline
$\times$ & $\times$ & $\times$ & 51.97 & 31.95 & 73.82 & 64.47 \\
$\times$ & $\times$ & $\checkmark$ & 54.64 & 34.71 & 76.38 & 66.99 \\
$\checkmark$ & $\times$ & $\times$ & 58.84 & 39.76 & 81.17 & 71.93 \\
$\checkmark$ & $\times$ & $\checkmark$ & 62.27 & 43.05 & 84.53 & 75.28 \\
$\checkmark$ & $\checkmark$ & $\times$ & 61.16 & 42.22 & 83.39 & 74.41 \\
\hline
$\checkmark$ & $\checkmark$ & $\checkmark$ & \textbf{64.34} & \textbf{45.68} & \textbf{86.92} & \textbf{77.75}
\\ \hline
\end{tabular}}}
\end{center}
\caption{Main ablation study on ActivityNet Captions dataset, where we remove each key individual component to investigate its effectiveness.}
\vspace{-10pt}
\label{tab:ablation1}
\end{table}

\noindent \textbf{Main ablation.}
To analyze how each model component contributes to the task, we perform main ablation study as shown in Table~\ref{tab:ablation1}.
We start from the baseline model which does not rely on both the tree-structure framework and the self-supervised strategy to address the one-shot TSL. Specifically, this baseline model directly generates multiple coarse segment proposals like previous supervised methods and then utilizes the rank loss in Eq.(\ref{eq:rank}) for training. It shows that this baseline achieves relatively worse performance compared to most weakly- and fully-supervised methods. By applying the self-supervised constraint of Eq.(\ref{eq:inter})(\ref{eq:intra}) to the baseline, the performance improves a lot. Applying the tree-structure framework brings the largest improvement since our well-designed merging strategy helps to distinguish the ambiguous adjacent frames for constructing more accurate segments under the limited supervision signals. Moreover, the self-supervised constraint seems to be more robust to the tree-structure framework due to the high quality of the segment hypotheses. The tree pruning strategy also helps the model to reduce the negative influence of the query-irrelevant frames. Overall, the whole tree-structure framework with both pruning strategy and self-supervised constraints achieves the best results.

\begin{table}[t!]
\small
\begin{center}
\scalebox{1.0}{
\setlength{\tabcolsep}{1.4mm}{
\begin{tabular}{c|c|cccc}
\hline
\multirow{2}*{Methods} & \multirow{2}*{Type} & R@1 & R@1 & R@5 & R@5\\
~ & ~ & IoU=0.3 & IoU=0.5 & IoU=0.3 & IoU=0.5\\
\hline\hline
2D-TAN & FS & 59.45 & 44.51 & 85.53 & 77.13 \\
2D-TAN* & OS & 48.76 & 34.29 & 75.88 & 66.02\\
\hline
WSTAN & WS & 52.45 & 30.01 & 79.38 & 63.42\\
WSTAN* & OS & 54.69 & 32.17 & 80.93 & 65.58 \\
\hline
\textbf{MHST} & OS & \textbf{64.34} & \textbf{45.68} & \textbf{86.92} & \textbf{77.75}
\\ \hline
\end{tabular}}}
\end{center}
\caption{Applying the one-shot setting to both fully- and weakly-supervised methods on ActivityNet Captions. }
\label{tab:ablation2}
\end{table}

\begin{table}[t!]
\small
\begin{center}
\scalebox{1.0}{
\setlength{\tabcolsep}{1.4mm}{
\begin{tabular}{cc|cccc}
\hline
Visual & Linguistic & R@1 & R@1 & R@5 & R@5\\
Similarity & Similarity & IoU=0.3 & IoU=0.5 & IoU=0.3 & IoU=0.5\\
\hline\hline
$\checkmark$ & $\times$ & 60.96 & 42.17 & 83.58 & 74.42 \\
$\times$ & $\checkmark$ & 62.10 & 43.34 & 84.66 & 75.25 \\
\hline
$\checkmark$ & $\checkmark$ & \textbf{64.34} & \textbf{45.68} & \textbf{86.92} & \textbf{77.75}
\\ \hline
\end{tabular}}}
\end{center}
\caption{Effect of visual- and linguistic similarities for node merging during the tree building.}
\label{tab:ablation3}
\end{table}

\begin{table}[t!]
\small
\begin{center}
\scalebox{1.0}{
\setlength{\tabcolsep}{1.4mm}{
\begin{tabular}{c|cccc}
\hline
\multirow{2}*{Hyperparameters} & R@1 & R@1 & R@5 & R@5\\
~ & IoU=0.3 & IoU=0.5 & IoU=0.3 & IoU=0.5\\
\hline\hline
$\alpha=50\%$ & 63.28 & 44.39 & 85.56 & 76.31 \\
$\alpha=60\%$ & \textbf{64.34} & 45.68 & \textbf{86.92} & \textbf{77.75} \\
$\alpha=70\%$ & 64.32 & \textbf{45.70} & 86.84 & 77.69
\\ \hline
$\tau=0.6$ & 63.47 & 44.82 & 85.83 & 76.95 \\
$\tau=0.7$ & \textbf{64.34} & \textbf{45.68} & \textbf{86.92} & \textbf{77.75} \\
$\tau=0.8$ & 63.96 & 45.33 & 86.60 & 77.41
\\ \hline
$L=1$ & 63.58 & 45.01 & 86.20 & 77.12 \\
$L=3$ & \textbf{64.34} & 45.68 & 86.92 & 77.75 \\
$L=5$ & 64.26 & \textbf{45.75} & \textbf{87.07} & \textbf{77.83}
\\ \hline
\end{tabular}}}
\end{center}
\caption{Effect of different hyperparameters for node merging during and tree pruning.}
\vspace{-10pt}
\label{tab:ablation4}
\end{table}

\begin{figure*}[t!]
\centering
\includegraphics[width=0.98\textwidth]{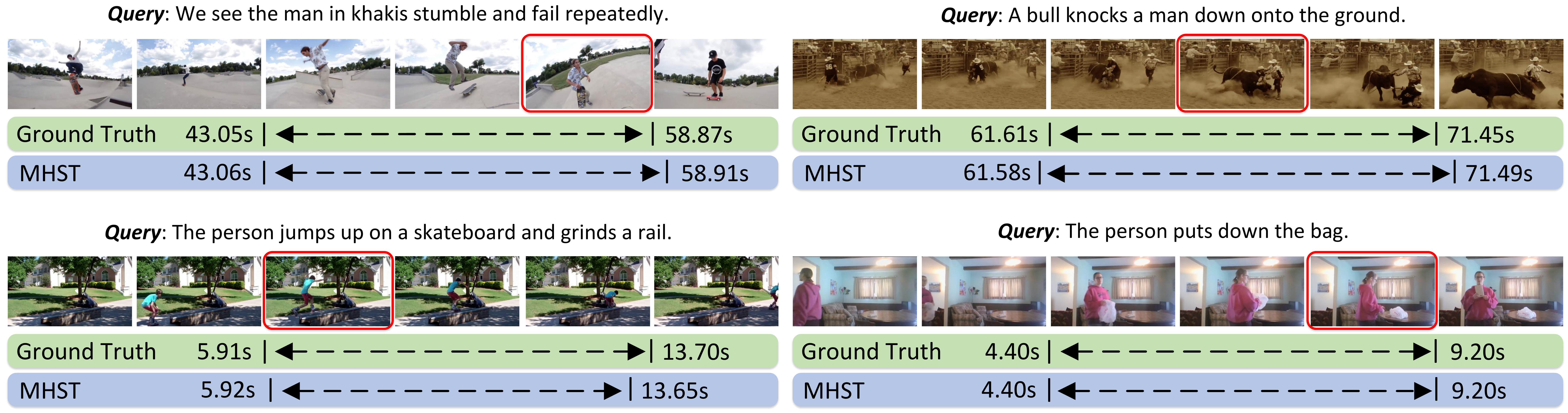}
\caption{Qualitative results on ActivityNet Captions and Charades-STA. The red rectangle denotes the labeled frame in video.}
\label{fig:result}
\vspace{-10pt}
\end{figure*}

\noindent \textbf{Effect of our one-shot pipeline.}
To fairly compare with the existing fully- and weakly-supervised methods, we re-implement some works with our one-shot setting as shown in Table~\ref{tab:ablation2}. Specifically, for 2D-TAN, we replace the loss of 2D map with our rank loss in Eq.(\ref{eq:rank}) for handling the weak labels. For WSTAN, we also add the rank loss for enriching the model contexts. We keep all the other settings being the same as their original works. From this table, we can find that, although one-shot setting brings additional one frame label to the weakly setting, the improvement of WSTAN* is still limited since it lacks sufficient self-supervision for discriminating the frame-wise representations and cannot generate accurate segment by measuring both visual-linguistic relevances like us. Moreover, the performance of 2D-TAN* degenerates a lot, demonstrating that previous fully setting is not robust to the weak label. Overall, it indicates that the one-shot setting is worth being investigated, and our proposed tree-structure framework is effectiveness.

\noindent \textbf{Effect of the visual-linguistic relevance.}
As shown in Table~\ref{tab:ablation3}, we investigate the effect of the visual- and linguistics relevances for node merging during the tree building process. It shows that both of them are crucial for segment construction, since the visual relevance helps to determine the visual similar adjacent frames near the labeled frame while the linguistic relevance helps to filter our the query-irrelevant frames.
By applying both of them for node merging, our model can achieve the best performance.

\noindent \textbf{Hyper-parameters of tree building and pruning.}
Moreover, we investigate the robustness of the proposed model to different hyper-parameters in Table~\ref{tab:ablation4}.
For the hyperparameter $\alpha$ in node merging, we find that the model achieves the best result when $\alpha$ is set to 60\%. Smaller $\alpha$ will lead negative node merging while larger $\alpha$ will filter out some positive nodes. As for $\tau$ in tree pruning, the model achieves the best result when $\tau=0.7$. Larger $\tau$ will filter out positive nodes. Besides, we also investigate the effect of different number $L$ in the pruning process. It shows that the model achieves the best result when $L=5$. However, lager $L$ gets the punishment in speed. Therefore, we set $L=3$ in all experiments.

\begin{table}[t!]
\small
\begin{center}
\scalebox{1.0}{
\setlength{\tabcolsep}{1.2mm}{
\begin{tabular}{ccc|cccc}
\hline
\multirow{2}*{$\mathcal{L}_{rank}$} & \multirow{2}*{$\mathcal{L}_{inter}$} & \multirow{2}*{$\mathcal{L}_{intra}$} & R@1 & R@1 & R@5 & R@5\\
~ & ~ & ~ & IoU=0.3 & IoU=0.5 & IoU=0.3 & IoU=0.5\\
\hline\hline
$\checkmark$ & $\times$ & $\times$ & 61.16 & 42.22 & 83.39 & 74.41 \\
$\checkmark$ & $\checkmark$ & $\times$ & 62.98 & 44.01 & 85.27 & 76.44 \\
$\checkmark$ & $\times$ & $\checkmark$ & 62.46 & 43.49 & 84.83 & 76.02 \\
$\checkmark$ & $\checkmark$ & $\checkmark$ & \textbf{64.34} & \textbf{45.68} & \textbf{86.92} & \textbf{77.75} 
\\ \hline
\end{tabular}}}
\end{center}
\caption{Ablation study on the supervision losses.}
\vspace{-10pt}
\label{tab:ablation5}
\end{table}

\noindent \textbf{Effect of the supervision losses.}
At last, we investigate the effectiveness of the proposed self-supervised losses of Eq.(\ref{eq:inter}) and (\ref{eq:intra}) in Table~\ref{tab:ablation5}.
Here, the ranking loss $\mathcal{L}_{rank}$ is the baseline for addressing the one-shot TSL setting.
It shows that the ranking loss $\mathcal{L}_{rank}$ is well-designed to the one-shot setting, and achieves great performance. Moreover, by applying the inter-constraint loss or the intra-constraint loss to the baseline, the model has significant performance by learning more discriminative frame-wise representations.
Overall, both two self-constraint losses $\mathcal{L}_{inter}$ and $\mathcal{L}_{intra}$ contributes a lot to the final performance, and we can achieve the best performance by jointly utilizing them.

\subsection{Qualitative Results}
As shown in Figure~\ref{fig:result}, we give the visualization of the localization results on both ActivityNet Captions and Charades-STA datasets for examples. In this figure, each video has only one frame (in red rectangle) labeled according to the query. It shows that our proposed MHST is robust to the one-shot setting and can well predict the segment boundaries.

\section{Conclusion}
In this paper, we \red{address the challenging} one-shot setting into the temporal sentence localization task. To achieve this goal, the model needs to make full use of the solely one labeled frame in each video to retrieve the target segment according to the query semantics.
Considering that the invalid frames unrelated to the query sentence or the labeled frame may bring confounding to the one-shot training process, we design a novel tree-structure framework called Multiple Hypotheses Segment Tree (MHST) to avoid this issue. 
Specifically, the hypotheses tree module merges adjacent frames sharing the similar visual-linguistic semantics
into a new upper node. Then, by iteratively building the tree and pruning the invalid nodes, we can get the complete and query-related root nodes which represent the segment hypotheses constructed by their contained consecutive frames. Finally, we develop several self-supervised losses to train the segment tree and predict the confidence scores for each segment hypothesis. During the inference, we directly choose the segment with highest score as the final prediction. Experimental results on two challenging datasets (ActivityNet Captions and Charades-STA) demonstrate that our MHST achieves a competitive performance compared to existing fully- and weakly-supervised methods.

\noindent \textbf{Acknowledgements.}
This work is supported by National Natural Science Foundation of China (NSFC) under Grant No. 61972448.

\bibliography{aaai22}

\end{document}